\documentclass[letterpaper]{article} 
\usepackage[draft]{aaai2026}
\usepackage{times}  
\usepackage{helvet}  
\usepackage{courier}  
\usepackage[hyphens]{url}  
\usepackage{graphicx} 
\usepackage{natbib}  
\usepackage{caption} 
\usepackage{algorithm}
\usepackage{algpseudocode}

\usepackage{newfloat}
\usepackage{listings}
\DeclareCaptionStyle{ruled}{labelfont=normalfont,labelsep=colon,strut=off} 
\floatstyle{ruled}
\newfloat{listing}{tb}{lst}{}
\floatname{listing}{Listing}
\usepackage{amsmath}
\usepackage{amssymb}
\usepackage{xspace}
\usepackage{pifont}
\newcommand{\ourmethod}{{\fontfamily{lmtt}\selectfont \textbf{Meta-R1}}\xspace}
\newcommand{\promethod}{{\fontfamily{lmtt}\selectfont \textbf{Meta-R1 Pro}}\xspace}
\usepackage[table,xcdraw,dvipsnames]{xcolor}
\definecolor{mc}{RGB}{228,219,232}
\definecolor{oc}{RGB}{255,255,224}
\usepackage{fontawesome5}
\usepackage{booktabs}
\usepackage{makecell}
\usepackage{bbding}
\usepackage{multirow}

\usepackage{enumitem}
\usepackage{pdfpages}

\title{\ourmethod: Empowering Large Reasoning Models with Metacognition}
\author{
    Haonan Dong\textsuperscript{\rm 1},
    Haoran Ye\textsuperscript{\rm 1},
    Wenhao Zhu\textsuperscript{\rm 1},
    Kehan Jiang\textsuperscript{\rm 1},
    Guojie Song\textsuperscript{\rm 1}\thanks{Corresponding author.}
}
\affiliations{
    \textsuperscript{\rm 1}State Key Laboratory of General Artificial Intelligence, \\ 
    School of Intelligence Science and Technology, Peking University\\
    {\faEnvelope} \texttt{hndong25@stu.pku.edu.cn, gjsong@pku.edu.cn}
}

\begin{document}

\maketitle

\begin{abstract}

Large Reasoning Models (LRMs) demonstrate remarkable capabilities on complex tasks, exhibiting emergent, human-like thinking patterns.
Despite their advances, we identify a fundamental limitation: current LRMs lack a dedicated meta-level cognitive system—an essential faculty in human cognition that enables “thinking about thinking”. This absence leaves their emergent abilities \textit{uncontrollable} (non-adaptive reasoning), \textit{unreliable} (intermediate error), and \textit{inflexible} (lack of a clear methodology). 
To address this gap, we introduce \ourmethod, a systematic and generic framework that endows LRMs with explicit metacognitive capabilities. Drawing on principles from cognitive science, \ourmethod decomposes the reasoning process into distinct \textit{object-level} and \textit{meta-level} components, orchestrating proactive planning, online regulation, and adaptive early stopping within a cascaded framework.
Experiments on three challenging benchmarks and against eight competitive baselines demonstrate that \ourmethod is: (\textbf{I}) \textbf{high-performing}, surpassing state-of-the-art methods by up to $27.3\%$; (\textbf{II}) \textbf{token-efficient}, reducing token consumption to $15.7\%\sim32.7\%$ and improving efficiency by up to $14.8\%$ when compared to its vanilla counterparts; and (\textbf{III}) \textbf{transferable}, maintaining robust performance across datasets and model backbones.

\end{abstract}

\section{Introduction}

\begin{figure}[!t]
\centering
\includegraphics[width=1\linewidth]{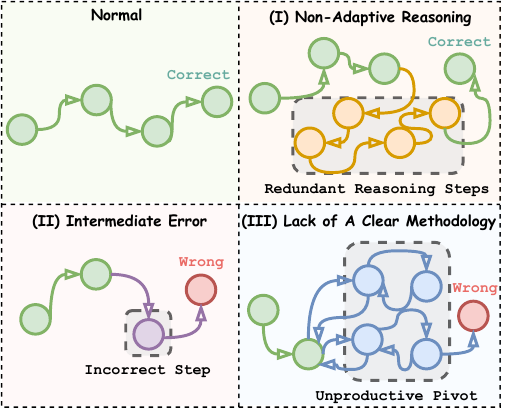} 
\caption{
    Three manifestations of metacognitive deficiency in existing LRMs.
    }  
\label{fig:intro}
\end{figure}
Large Reasoning Models (LRMs), exemplified by OpenAI-o1 \cite{openai-o1} and DeepSeek-R1 \cite{deepseek-r1}, have achieved substantial progress across a wide range of complex tasks \cite{survey-reasoning-1,survey-reasoning-2}. Upon receiving a query, these models explicitly generate step-by-step chains of thought (CoT) \cite{cot} prior to producing final answers. They exhibit the capacity for \textit{test-time scaling}, yielding improved solutions and tackling more challenging problems by extending reasoning duration \cite{test-time-1,test-time-2,test-time-3}. Glimpses of their reasoning reveal emergent, human-like cognitive behaviours—including \textit{self-reflection}, \textit{backtracking}, and \textit{self-verification} \cite{rl-1,r1-test}—often attributed to reinforcement learning (RL) \cite{rl-2,rl-3}. Recent advances have further enhanced these models by increasing data efficiency \cite{data-1,data-2,test-time-3} and introducing novel training algorithms \cite{novel-rl-1,novel-rl-2,s-grpo}.

However, closer examinations of the mainstream LRMs \cite{deepseek-r1} reveal that their emergent behaviors are intrinsically uncontrollable, unreliable, and inflexible. As depicted in Figure \ref{fig:intro}, these models frequently produce redundant, erroneous, or unproductive reasoning steps or pivots. 
At a \textbf{surface level}, LRMs (\textbf{I}) fail to adaptively regulate the length of their reasoning in accordance with problem complexity, resulting in substantial token inefficiency; (\textbf{II}) exhibit persistent procedural errors, such as miscalculations in complex tasks; and (\textbf{III}) display insufficient methodological awareness, often manifesting as overconfidence, superficial attempts, or frequent, unwarranted changes in strategy. Indeed, similar findings are presented by \cite{overthinking,cot-detect,cot-explain,cot-faithful}. At a \textbf{deeper level}, these deficiencies arise from the models' autoregressive architecture, which impedes effective self-monitoring and regulation during next-token generation \cite{deeper-level-1,deeper-level-2}.

These deficiencies collectively reveal a fundamental lack of metacognition in LRMs—a core human cognitive faculty underpinning the ability to ``think about thinking'' \cite{metacog-decades-1,metacog-decades-2}. Metacognition orchestrates \textit{strategic resource allocation}, \textit{error detection and correction}, and \textit{the flexible adaptation of reasoning processes}, all of which are essential for advanced human cognition \cite{metacog-1,metacog-2,metacog-3}. According to the well-established two-level model of Nelson and Narens \cite{metamemory}, metacognition comprises a meta-level, which governs the planning and regulation of reasoning, and an object-level, which executes the reasoning itself. Neurocognitive evidence further supports the existence of specialized brain regions dedicated to metacognitive control \cite{meta-neural}. In contrast, current monolithic autoregressive architectures and RL training regimes lack the structural and functional complexity required to elicit such sophisticated, brain-like reasoning. We therefore argue that explicit cognition engineering \citep{cog-eng} is essential to scaffold LRMs toward their full potential.

In response to these challenges, we present \ourmethod, a systematic and generic framework that confers explicit metacognitive capabilities upon LRMs. Drawing on principles from cognitive science, \ourmethod decomposes the reasoning process into distinct \textit{object-level} and \textit{meta-level} modules, orchestrating three core metacognitive functions: \ding{182} \textbf{Proactive Metacognitive Planning}, which formalizes problem definitions, assesses difficulty, and selects strategies accordingly; \ding{183} \textbf{Online Metacognitive Regulation}, which facilitates real-time monitoring and control between levels through a pre-defined communication protocol; and \ding{184} \textbf{Satisficing Termination}, which determines when to conclude reasoning.
Through extensive experiments on canonical mathematical benchmarks (MATH500, AIME24, GSM8K) against eight competitive baselines, we demonstrate that \ourmethod is (\textbf{I}) \textbf{high-performing}, surpassing state-of-the-art (SOTA) methods by up to $27.3\%$; (\textbf{II}) \textbf{token-efficient}, reducing token consumption to $15.7\%\sim32.7\%$ and improving efficiency by up to $14.8\%$ compared to its vanilla counterparts; and (\textbf{III}) \textbf{transferable}, maintaining robust performance across datasets and model backbones. Our main contributions are summarized as follows:

\begin{itemize}[leftmargin=*]
\item[\ding{96}] \emph{\textbf{Paradigm Shift.}} We identify the absence of metacognition as a fundamental limitation in current LRMs. By introducing metacognitive theory into the reasoning process, we advance large language model (LLM) reasoning towards the paradigm of explicit cognition engineering.

\item[\ding{96}] \emph{\textbf{Practical Framework.}} We present \ourmethod, a systematic framework that decouples the metacognitive system into distinct object- and meta-levels, and implements a systematic three-stage workflow for task resolution.

\item[\ding{96}] \emph{\textbf{Experimental Validation.}} Extensive experiments on canonical benchmarks demonstrate that \ourmethod (\textbf{I}) outperforms baseline methods by up to 27.3\%, (\textbf{II}) reduces token consumption to $15.7\% \sim 32.7\%$ of that required by vanilla models, and (\textbf{III}) effectively transfers across datasets and models.
\end{itemize}

\section{Method}
\begin{figure*}[!ht]
  \centering
  \includegraphics[width=\linewidth]{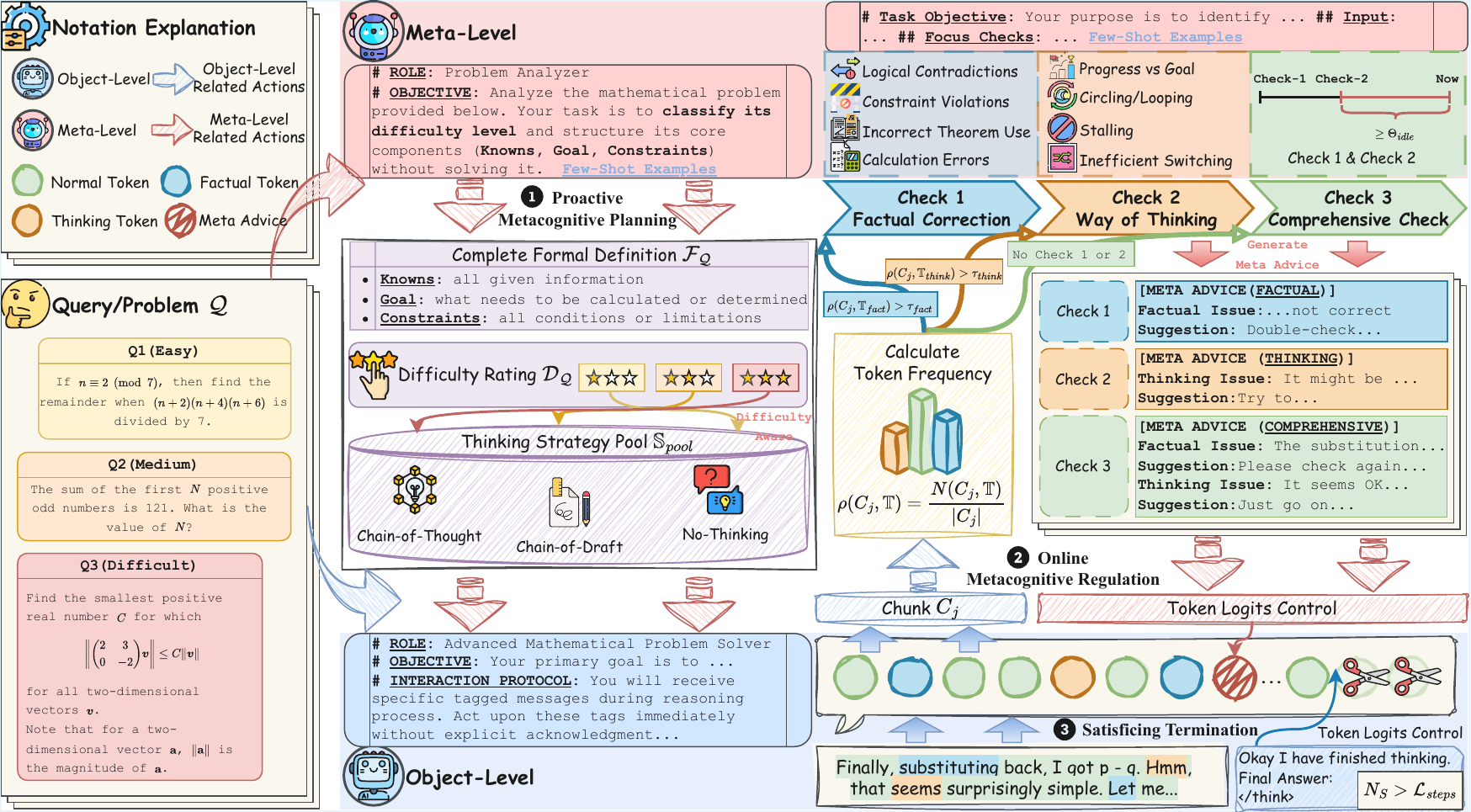}
  \caption{The methodological framework of \ourmethod. Given a query, the process involves: \ding{182} meta-level preparation, \ding{183} real-time monitoring and control between the two levels of reasoning, culminating in \ding{184} system-triggered early stopping.}
   \label{fig:overview}
\end{figure*}

Figure \ref{fig:overview} illustrates how \ourmethod augments existing LRMs with metacognitive capabilities. Building on the theoretical framework of Nelson and Narens \cite{metamemory}, our approach conceptualizes metacognition as a two-level architecture: an \textit{object-level}, which directly engages with the task, and a \textit{meta-level}, which regulates the object-level. In this design, the main LRM naturally serves as the object-level, while an additional LLM is introduced to function as the meta-level. \ourmethod implements a comprehensive three-stage process for task resolution. Specifically, given a query, the workflow proceeds as follows: \ding{182} \textbf{Proactive Metacognitive Planning}: Before problem-solving begins, the meta-level conducts an in-depth analysis of the problem to allocate cognitive resources. \ding{183} \textbf{Online Metacognitive Regulation}:
During problem-solving, information flows continuously between the meta-level and object-level through dynamic \textit{monitoring} and \textit{control} mechanisms, mediated by a pre-defined communication protocol, enabling real-time metacognitive regulation. \ding{184} \textbf{Satisficing Termination}: The framework determines the appropriate moment to conclude the reasoning process and generates the final response.

\subsection{Proactive Metacognitive Planning}
From a cognitive science perspective, the primary distinction between experts and novices lies in their preparatory phase of problem-solving \cite{expert-novice-1,expert-novice-2}. Experts, guided by metacognition, devote substantial cognitive effort to understanding the problem's intricacies and assessing its complexity before allocating resources. In contrast, novices often bypass this critical analysis, prematurely resorting to trial-and-error or rote application of formulas. Our framework operationalizes this phase at the meta-level through a structured procedure: (\textbf{I}) Schema Activation for Problem Formalization, (\textbf{II}) Ease-of-Learning Judgments for Difficulty Assessment, and (\textbf{III}) Cognitive Resource Allocation via Strategy Selection.

\paragraph{Schema Activation for Problem Formalization.} Schema activation entails retrieving relevant structured knowledge from long-term memory to interpret new problems \cite{schema-1,schema-2,schema-3}. This process facilitates effective reasoning by enabling recognition of a problem's underlying logical structure beyond its superficial features.
Accordingly, upon receiving a query $\mathcal{Q}$, the meta-level is prompted via few-shot learning to perform a formal problem decomposition based on a context $\mathcal{E}_{fp}$, yielding a structured problem definition $\mathcal{F}_{\mathcal{Q}}$. This definition is formulated as a tuple comprising three elements: \textit{Knowns} (the explicit givens and premises) $\mathcal{K}_{\mathcal{Q}}$, the \textit{Goal} (the target outcome to be found, computed, or proven) $\mathcal{G}_{\mathcal{Q}}$, and \textit{Constraints} (the explicit or implicit conditions that govern the problem space) $\mathcal{C}_{\mathcal{Q}}$. We formalize this decomposition as follows:
\begin{equation}
\mathcal{F}_{\mathcal{Q}} \leftarrow \mathcal{M}_{meta}(\mathcal{Q}\, | \, \mathcal{E}_{fp})
\label{eq:scema-activation}
\end{equation}
\begin{equation}
\mathcal{F}_{\mathcal{Q}} = (\mathcal{K}_{\mathcal{Q}}, \mathcal{G}_{\mathcal{Q}}, \mathcal{C}_{\mathcal{Q}})
\end{equation}

\paragraph{Ease-of-Learning Judgments for Difficulty Assessment.} Before tackling a cognitive task, individuals make metacognitive predictions about its difficulty, known as Ease-of-Learning (EOL) judgments \cite{metamemory,eol-1,eol-2}. We operationalize this approach by categorizing problem difficulty into three broad levels: \textit{Hard}, \textit{Medium}, and \textit{Easy}. Using few-shot learning, the meta-level classifies the difficulty $\mathcal{D}_{\mathcal{Q}}$ of a given problem $\mathcal{Q}$ based on a context $\mathcal{E}_{fp}$. While solving the problem may be complex, assessing its difficulty is comparatively straightforward. Formally, this process is defined as:
\begin{equation}
\mathcal{D}_{\mathcal{Q}} \leftarrow \mathcal{M}_{meta}(\mathcal{Q}\, | \, \mathcal{E}_{fp})
\label{eq:eol}
\end{equation}

\paragraph{Cognitive Resource Allocation via Strategy Selection.} Faced with multiple tasks or complex problems, individuals must efficiently allocate limited cognitive resources \cite{allocate-1,allocate-2,allocate-3}. This cognitive resource allocation, guided by prior difficulty assessment, determines the appropriate level of effort. In \ourmethod, we first construct a strategy pool $\mathbb{S}_{pool}$ of mainstream reasoning strategies. Based on the meta-level's difficulty assessment, we select a suitable reasoning strategy $\mathcal{S}_{thinking}$ for the object-level. This strategy is then combined with the formal problem definition $\mathcal{F}_{\mathcal{Q}}$ to construct the initial prompt for the object-level.

\subsection{Online Metacognitive Regulation}
After proactive metacognitive planning, the process advances to core problem-solving. According to Nelson and Narens' theory \cite{metamemory}, this stage involves two main activities: monitoring and control. Information flows between the object-level and meta-level in a bidirectional manner—monitoring sends information from the object-level to the meta-level, while control sends it in the opposite direction. Metacognitive regulation, or "thinking about thinking," focuses on detecting two error types: (\textbf{I}) factual errors, such as mistakes in specific solution steps, and (\textbf{II}) thinking errors, which are flaws in the reasoning methodology itself.

\begin{algorithm}[t]
\caption{\ourmethod Inference}
\label{alg:ourmethod}
\begin{algorithmic}[1]
\Require Query $\mathcal{Q}$, Meta-level $\mathcal{M}_{\text{meta}}$, Object-level $\mathcal{M}_{\text{obj}}$
\Ensure Final Answer $\mathcal{A}_{\text{final}}$

\Statex \textbf{Stage 1: Proactive Metacognitive Planning}
\Statex \textit{// Problem formalization}
\State $\mathcal{F}_{\mathcal{Q}} \leftarrow \mathcal{M}_{\text{meta}}(\mathcal{Q}\, | \, \mathcal{E}_{fp})$ \Comment{Eq. \ref{eq:scema-activation}}
\Statex \textit{// Difficulty Assessment}
\State $\mathcal{D}_{\mathcal{Q}} \leftarrow \mathcal{M}_{\text{meta}}(\mathcal{Q}\, | \, \mathcal{E}_{fp})$ \Comment{Eq. \ref{eq:eol}}
\State Select strategy $\mathcal{S}_{\text{thinking}}$ and step budget $B$ based on $\mathcal{D}_{\mathcal{Q}}$
\State $Prompt_{\text{init}} \leftarrow \text{ConstructPrompt}(\mathcal{F}_{\mathcal{Q}}, \mathcal{S}_{\text{thinking}})$
\State Initialize ReasoningTrace $\mathcal{R} \leftarrow \emptyset$, $i \leftarrow 0$, $i_{\text{last}} \leftarrow 0$

\Statex \textbf{Stage 2: Online Metacognitive Regulation}
\While{$i < B$} \Comment{Loop until budget is exceeded}
    \Statex \textit{// Object-level generates the next reasoning chunk}
    \State $C_i \leftarrow \mathcal{M}_{\text{obj}}(Prompt_{\text{init}} \cup \mathcal{R})$
    \State $\mathcal{R} \leftarrow \mathcal{R} \cup C_i$; $i \leftarrow i + 1$
    \Statex \textit{// Meta-level monitors for potential errors}
    \State Trigger control action $A_i$ based on token frequency anomaly or periodic safety interval \Comment{Eq. \ref{eq:control_trigger}}
    \If{$A_i \neq \emptyset$} 
        \State Get error type $E_{\text{type}}$ from action $A_i$
        \Statex \textit{// Meta-level generates advice upon detecting an error}
        \State $\mathcal{A}_{\text{meta}} \leftarrow \mathcal{M}_{\text{meta}}(C_i, E_{\text{type}} \, | \, \mathcal{E}_{fc})$ \Comment{Eq. \ref{eq:control}}
        \If{$\mathcal{A}_{\text{meta}}$ is not empty}
            \Statex \textit{\quad // Inject advice into the reasoning trace}
            \State $\mathcal{R} \leftarrow \mathcal{R} \cup \mathcal{A}_{\text{meta}}$ \Comment{Eq. \ref{eq:latent_prompt}}
        \EndIf
        \State $i_{\text{last}} \leftarrow i$
    \EndIf
\EndWhile

\Statex \textbf{Stage 3: Satisficing Termination}
\Statex \textit{// Force termination and generate the final answer}
\State Inject termination latent prompt into $\mathcal{M}_{\text{obj}}$
\State $\mathcal{A}_{\text{final}} \leftarrow \mathcal{M}_{\text{obj}}(Prompt_{\text{init}} \cup \mathcal{R})$
\State \Return $\mathcal{A}_{\text{final}}$
\end{algorithmic}
\end{algorithm}

\paragraph{Monitoring.} To efficiently detect and classify errors, we analyze the tokens generated during reasoning. Similar to prior works \cite{token-prior-1,token-prior-2,token-prior-3}, we empirically observe that R1-like models generate a series of specific keywords during their reasoning process, such as ``let'' and ``assume'' during mathematical derivations, or ``wait'' and ``alternatively'' when shifting strategies. Specifically, tokens are grouped into two categories: \textit{factual tokens} and \textit{thinking tokens}, corresponding to factual and thinking errors, respectively (see Appendix C). For analysis, we define a \textit{step} as a single paragraph of output, and a \textit{chunk} as a sequence of steps. After generating chunk $C_i$, we compute the frequency of each token type $\mathbb{T}$:
\begin{equation}
\rho(C_i, \mathbb{T}) = \frac{N(C_i, \mathbb{T})}{|C_i|}
\end{equation}
where $|C_i|$ is the total number of tokens in chunk $C_i$, and $N(C_i, \mathbb{T})$ counts the occurrences of tokens from set $\mathbb{T}$. Our monitoring mechanism uses two triggers: an \textit{anomaly-based trigger} and a \textit{periodic safety trigger}. The anomaly-based trigger is threshold-based: if the frequency of any token type exceeds a predefined threshold $\tau$, the corresponding error type is logged and control action $\mathcal{V}$ is triggered. The periodic safety trigger provides regular oversight by defining a \textit{safety interval} $\Theta_{safe}$, the maximum number of chunks allowed without a trigger. If no monitoring occurs within this interval, a comprehensive control action $\mathcal{V}_{\text{all}}$ for both factual and thinking errors is triggered. Formally, the action $A_i$ triggered for chunk $C_i$ is determined as follows:
\begin{equation}
\label{eq:control_trigger}
A_i =
\begin{cases}
\mathcal{V}_{\text{fact}} & \text{if } \rho(C_i, \mathbb{T}_{\text{fact}}) > \tau_{\text{fact}} \\
\mathcal{V}_{\text{think}} & \text{if } \rho(C_i, \mathbb{T}_{\text{think}}) > \tau_{\text{think}} \\
\mathcal{V}_{\text{all}} & \text{if } i - i_{\text{last}} > \Theta_{safe} \\
\emptyset & \text{otherwise}
\end{cases}
\end{equation}

\paragraph{Control.} Upon receiving a chunk flagged for a potential error, the meta-level initiates a targeted check for the corresponding error type $E_{\text{type}}$, using a dedicated few-shot prompt $\mathcal{E}_{fc}$. To identify factual errors, the meta-level checks four aspects: (I) internal logical contradictions within the current chunk; (II) violations of the constraints $\mathcal{C}_{\mathcal{Q}}$ established in the formal problem definition; (III) incorrect theorem application, specifically whether the necessary preconditions are met; and (IV) computational errors. To identify thinking errors, the meta-level assesses four conditions: (I) lack of progress towards the problem's goal $\mathcal{G}_{\mathcal{Q}}$; (II) reasoning loops, where the current line of thought is stagnating or repetitive; (III) stalled progress, indicating that the reasoning process is stuck; and (IV) strategic instability, such as prematurely abandoning a reasoning chain or frequently switching between approaches. If the meta-level confirms an error, it generates \texttt{META ADVICE}, denoted as $\mathcal{A}_{\text{meta}}$, which details the specific manifestation $e_\text{spec}$ of the error and provides actionable suggestions $s_\text{act}$ for correction. Finally, $\mathcal{A}_{\text{meta}}$ is passed as an information stream to the object-level. We formalize this process as follows:
\begin{equation}
\label{eq:control}
\mathcal{A}_{\text{meta}} \leftarrow\mathcal{M}_{meta}(C_i, E_{\text{type}} \, | \, \mathcal{E}_{fc})
\end{equation}
\begin{equation}
\label{eq:advice-generation}
\mathcal{A}_{\text{meta}} = (e_{\text{spec}}, s_{\text{act}})
\end{equation}

\paragraph{Communication Protocols.} We define our communication protocols for monitoring and control actions to enable information flow between the object-level and the meta-level. For a monitoring action, we explicitly provide the flagged chunk and its suspected error type to the meta-level as part of a standard prompt. For a control action, we introduce the \textit{latent prompt}. In contrast to conventional prompt engineering, this method achieves dynamic, real-time intervention through token-level control. Specifically, we consider the object-level $\mathcal{M}_{obj}$, as a standard autoregressive generator. At each generation step $t$, given the preceding sequence of tokens $x_{<t}$, it outputs a probability distribution $P_t$ for the next token $x_t$:
\begin{equation}
    P_t(x_t \, | \, x_{<t}) = \text{Softmax}(\mathcal{M}_{obj}(x_{<t}))
    \label{eq:standard_generation}
\end{equation}
The latent prompt method intervenes in this process to inject the \texttt{META ADVICE}, denoted as a token sequence $\mathcal{A}_{\text{meta}} = (a_1, a_2, \dots, a_L) \text{ of length } L$. This injection occurs over $L$ consecutive generation steps, starting from step $t_{\text{start}}$. For each step $t$ during this injection (where $t = t_{\text{start}} + k - 1 \text{ for } k=1, \dots, L$), we override the original distribution $P_t$ with a modified distribution $P'_t$. This new distribution forces the generation of the specific advice token $a_k$ by setting its probability to 1, effectively creating a Dirac delta distribution centered on $a_k$:
\begin{equation}
    \label{eq:latent_prompt}
    P'_{t_{\text{start}}+k-1}(x \, | \, x_{<t_{\text{start}}}, a_{<k}) =
    \begin{cases}
        1 & \text{if } x = a_k \\
        0 & \text{otherwise}
    \end{cases}
\end{equation}
The viability of this dynamic injection hinges on a predefined \texttt{INTERACTION PROTOCOL} within the object-level's system prompt, which familiarizes the object-level with the nature of the advice.

\begin{table*}[t]
  \centering
  \small
  \setlength{\tabcolsep}{1mm}
  \renewcommand{\arraystretch}{1}
    \begin{tabular}{l *{12}{c}}
      \toprule
      \multirow{2}{*}{\textbf{Method}}
        & \multicolumn{3}{c}{\textbf{GSM8K}}
        & \multicolumn{3}{c}{\textbf{AIME2024}}
        & \multicolumn{3}{c}{\textbf{MATH500}}
        & \multicolumn{3}{c}{\textbf{Avg.}} \\
      \cmidrule(lr){2-4}\cmidrule(lr){5-7}\cmidrule(lr){8-10}\cmidrule(lr){11-13}
        & Acc\,$\uparrow$  & Tokens\,$\downarrow$ & RSE\,$\uparrow$
        & Acc\,$\uparrow$  & Tokens\,$\downarrow$ & RSE\,$\uparrow$
        & Acc\,$\uparrow$  & Tokens\,$\downarrow$ & RSE\,$\uparrow$
        & Acc\,$\uparrow$  & Tokens\,$\downarrow$ & RSE\,$\uparrow$ \\
      \midrule

      \multicolumn{13}{c}{\textit{\textbf{DeepSeek-R1-Distill-Qwen-14B}}} \\
      Vanilla             & 94.2 & 2129 & 88.6 & 64.4 & 11099 & 50.0 & 93.5 & 3844 & 84.2 & 84.0 & 5691 & 74.3 \\
      DEER                & 93.3 & 982  & 90.6 & \textbf{70.0} & 10335 & \underline{54.8} & 91.4 & 2753 & 84.6 & 84.9 & 4690 & 76.7 \\
      GRPO                & 95.3 & 2120 & 89.7 & 65.8 & 13504 & 48.7 & 84.0 & 4471 & 74.5 & 81.7 & 6698 & 71.0 \\
      RL + Length Penalty & 94.7 & 775  & 92.5 & 55.0 & 7950  & 45.1 & 92.4 & \textbf{1993} & \underline{87.3} & 80.7 & \underline{3573} & 75.0 \\
      S-GRPO              & \textbf{96.2} & 724  & \underline{94.1} & 64.4 & \textbf{6712} & 54.2 & 93.6 & \underline{2146} & \textbf{88.0} & 84.7 & \textbf{3194} & \underline{78.8} \\
      \ourmethod          & \underline{95.9} & \textbf{691}  & 93.9 & \underline{66.7} & \underline{7899} & \underline{54.8} & \underline{94.5} & 2903 & 87.1 & \underline{85.7} & 3831 & 78.6 \\
      \promethod          & \textbf{96.2} & \underline{717} & \textbf{94.2} & \textbf{70.0} & 8020 & \textbf{57.4} & \textbf{94.8} & 3085 & 87.0 & \textbf{87.0} & 3941 & \textbf{79.5} \\
      \midrule

      \multicolumn{13}{c}{\textit{\textbf{DeepSeek-R1-Distill-Qwen-32B}}} \\
      Vanilla             & 95.6 & 875  & 93.2 & 72.0 & 9347 & 57.5 & 94.5 & 3543 & 85.7 & 87.4 & 4588 & 78.8 \\
      DEER                & 95.1 & 819  & 92.8 & 63.3 & 7424 & 52.5 & 90.4 & \textbf{2425} & 84.4 & 82.9 & 3556 & 76.6 \\
      Greedy              & 95.3 & 1048 & 92.4 & 63.3 & 8050 & 51.8 & 93.0 & 3651 & 84.1 & 83.9 & 4250 & 76.1 \\
      Soft Thinking       & 95.8 & 785  & \underline{93.6} & \textbf{76.7} & \textbf{6620} & \textbf{64.7} & \underline{95.0} & 3373 & 86.5 & \underline{89.2} & 3593 & \underline{81.6} \\
      \ourmethod          & \underline{96.3} & \textbf{651} & \textbf{94.4} & \underline{73.3} & \underline{6720} & 61.7 & \underline{95.0} & \underline{2535} & \textbf{88.4} & 88.2 & \textbf{3302} & 81.5 \\
      \promethod          & \textbf{96.5} & \underline{733} & \textbf{94.4} & \textbf{76.7} & 6978 & \underline{64.2} & \textbf{95.2} & 2722 & \underline{88.2} & \textbf{89.5} & \underline{3478} & \textbf{82.3} \\
      \bottomrule
    \end{tabular}
  \caption{Performance comparison of \ourmethod versus baselines on the GSM8K, AIME2024, and MATH500 benchmarks, using DeepSeek-R1-Distill-Qwen-14B and 32B. We report accuracy (Acc), token consumption (Tokens), and efficiency (RSE). Higher values are better for Acc and RSE, lower values are better for Tokens. The best results are highlighted in \textbf{bold}, and the runners-up are \underline{underlined}.}
   \label{tab:main}
\end{table*}

\subsection{Satisficing Termination}
To ensure both efficiency and plausibility in reasoning, we introduce a Satisficing Termination mechanism, which is grounded in Herbert Simon's work on Bounded Rationality and Satisficing \cite{terminate-1,terminate-2,terminate-3}. The theory states that individuals do not seek to optimize, but rather to find a good enough solution and then conclude their search. To apply this principle, we set distinct reasoning step budgets for problems of different difficulty levels. If the object-level exceeds this budget, we intervene with our latent prompt method (``\texttt{\textbackslash n\textbackslash nOkay I have finished thinking.\textbackslash nFinal Answer: \textbackslash n</think>\textbackslash n}'') to force the immediate termination of the reasoning process, prompting the object-level to synthesize and output the best possible answer based on its existing line of thought.

\section{Experiments}
In this section, we conduct extensive experiments to answer the following research questions: ($\boldsymbol{\mathcal{RQ}1}$) How does \ourmethod perform across various model scales and datasets? ($\boldsymbol{\mathcal{RQ}2}$) How token-efficient is \ourmethod? ($\boldsymbol{\mathcal{RQ}3}$) How does the scale of meta-level affect both performance and token efficiency? ($\boldsymbol{\mathcal{RQ}4}$) Can meta-level accurately assess problem difficulty? ($\boldsymbol{\mathcal{RQ}5}$) What is the contribution of each stage?
\subsection{Experimental Setup}
\paragraph{Baselines.} For our baseline evaluation, we select a range of representative baseline methods, which are categorized into three groups: $\blacksquare$ \textbf{Vanilla Models}: DeepSeek-R1-Distill-Qwen-14B and DeepSeek-R1-Distill-Qwen-32B \cite{deepseek-r1}; $\blacksquare$ \textbf{Training-Free Methods}: DEER \cite{deer}, Greedy CoT Thinking \cite{cot}, and Soft Thinking \cite{soft}; and $\blacksquare$ \textbf{RL-based Methods}: GRPO \cite{grpo}, RL + Length Penalty \cite{rl+lp}, and S-GRPO \cite{s-grpo}.
\paragraph{Datasets.} We evaluate \ourmethod on three benchmarks widely used for assessing advanced reasoning in large models: GSM8K \cite{gsm8k}, MATH500 \cite{math500}, and AIME2024 \cite{aime}. Detailed information about the datasets is provided in Appendix B.
\paragraph{Metrics.} For our evaluation metrics, we not only consider accuracy and total token consumption, but also aim to comprehensively assess the model's reasoning efficiency. Prior works have often measured efficiency using linear metrics \cite{cothink,efficient-thinking-survey}, which can overlook the diminishing marginal returns of token consumption. To address this limitation, we propose a new metric: \textit{\textbf{Root-Scaled Efficiency}} (RSE). Specifically, it is defined as:
\begin{equation}
\mathrm{RSE}
= \frac{A}{\sqrt{1 + \frac{L}{L_{\max}}}},
\end{equation}
where $A$ denotes the accuracy, $L$ represents the total token consumption, and $L_{\max}$ is the maximum output length of the model in tokens. See Appendix D for more details.
\paragraph{Implementation Details.} For the object-level, we select two prominent large reasoning models: DeepSeek-R1-Distill-Qwen-14B and DeepSeek-R1-Distill-Qwen-32B \cite{deepseek-r1}. For the meta-level, we employ two small instruct models \cite{qwen25,qwen25-model}, defining our two framework variants: \ourmethod (using Qwen2.5-1.5B-Instruct) and \promethod (using Qwen2.5-3B-Instruct). All meta-level LLMs are accessed via API. Our \textit{strategy pool} is composed of three reasoning methods: Chain-of-Thought \cite{cot}, Chain-of-Draft \cite{cod}, and No-Thinking \cite{nothinking}. All experiments are conducted on four NVIDIA GeForce RTX 4090 GPUs. All parameter settings and prompts are detailed in Appendix E.

\subsection{Main Results ($\boldsymbol{\mathcal{RQ}1\,\&\,\mathcal{RQ}2}$)}
\begin{figure}[!t]
\centering
\includegraphics[width=1\linewidth]{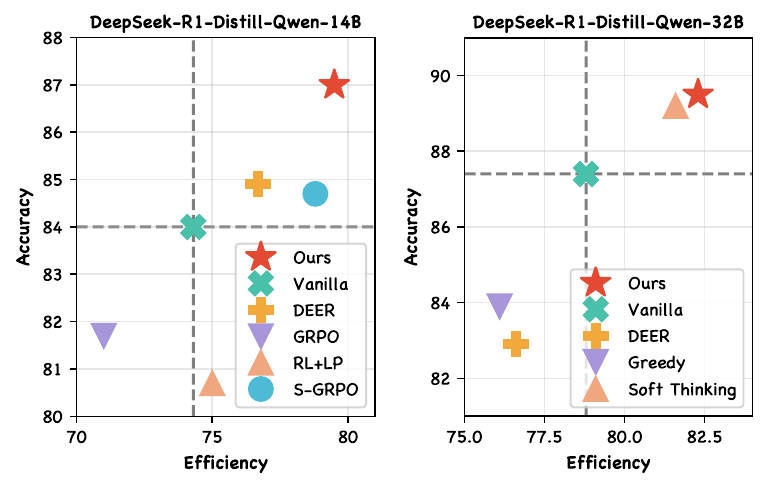} 
\caption{
    Comparison of accuracy and efficiency between \ourmethod and baseline methods on two models.
    }  
\label{fig:acc-eff}
\end{figure}

To answer $\mathcal{RQ}$1 and $\mathcal{RQ}$2, we conduct a comparative analysis of \ourmethod against eight baselines on three widely-used benchmarks: GSM8K, AIME2024, and MATH500. The evaluation focuses on accuracy, token consumption, and overall efficiency, with the results summarized in Table \ref{tab:main}. Based on these results, we have the following observations:
\paragraph{Obs. \ding{182} \ourmethod achieves optimal performance across all model scales and datasets.} On average, \ourmethod improves performance by $0.9\% \sim 3.6\%$ compared to the vanilla models. Against the other baseline methods, the average improvement is even more pronounced, ranging from $0.3\% \sim 8.0\%$, with a peak gain of $27.3\%$. Specifically, when using DeepSeek-R1-Distill-Qwen-14B as the backbone on AIME2024, \promethod surpasses the vanilla baseline by $8.7\%$. Similarly, with the DeepSeek-R1-Distill-Qwen-32B backbone on MATH500, \promethod achieves a $5.3\%$ improvement over DEER.
\paragraph{Obs. \ding{183} \ourmethod demonstrates significant token efficiency.} Compared to vanilla models, \ourmethod reduces token consumption by an average of $24.2\% \sim 32.7\%$. Notably, on GSM8K, \ourmethod ranks as the runner‑up in accuracy across all model scales while consistently using a minimal number of tokens. In terms of our comprehensive efficiency metric, \ourmethod attains the top RSE values. It boosts efficiency by an average of $3.4\% \sim 7.0\%$ over vanilla models, with a maximum increase of $14.8\%$. Specifically, on MATH500, \ourmethod shows a $16.9\%$ efficiency gain over GRPO. In Figure \ref{fig:acc-eff}, we plot the average performance versus efficiency of all methods, aggregated across two backbones and three benchmarks. The visualization clearly demonstrates that, compared to the baseline methods, \ourmethod achieves both better performance and higher efficiency. 

\subsection{Analysis of Meta-Level ($\boldsymbol{\mathcal{RQ}3\,\&\,\mathcal{RQ}4}$)}
\begin{figure}[!t]
\centering
\includegraphics[width=1\linewidth]{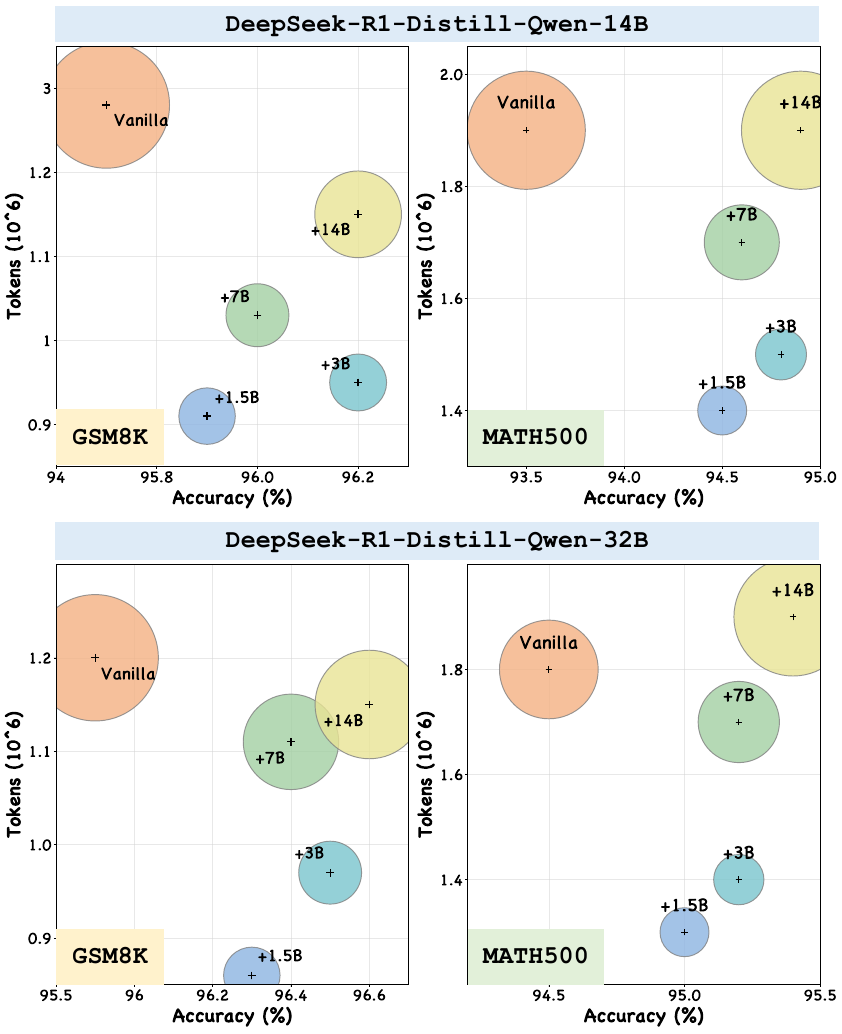} 
\caption{
    Effect of meta-level model scale on accuracy and token consumption. We vary the size of meta-level (1.5B to 14B) while using two fixed object-level models (DeepSeek-R1-Distill-Qwen-14B and 32B), with evaluations performed on the GSM8K and MATH500 benchmarks.
    }  
\label{fig:model-scale}
\end{figure}

Firstly, we posit that instruct models, in contrast to reasoning models, are characterized by concise and rapid responses \cite{cothink,speculative-thinking}. For this reason, we adopt them for the meta-level. To answer $\mathcal{RQ}$3 and $\mathcal{RQ}$4, we conduct a series of analytical experiments focusing on the meta-level, from which we derive the following observations: 
\paragraph{Obs. \ding{184} Small models suffice for an effective and efficient meta-level.} To compare the impact of using different model sizes for the meta-level, we test four Qwen2.5-Instruct models of varying sizes (1.5B, 3B, 7B, 14B) as the meta-level across our two backbones on GSM8K and MATH500. As shown in Figure \ref{fig:model-scale}, while scaling up the meta-level model size does yield a corresponding increase in both performance and token cost, we observe that the 1.5B and 3B models strike the optimal balance. For instance, with the DeepSeek-R1-Distill-Qwen-14B backbone, scaling the meta-level from 3B to 7B yields no performance improvement despite a $14.4\%\uparrow$ increase in token consumption, while scaling to 14B provides a marginal $0.1$ gain at the cost of $18.6\%\uparrow$ more tokens. For this reason, we adopt the 1.5B and 3B models as the meta-level in our main experiments.

\begin{figure}[!t]
\centering
\includegraphics[width=1\linewidth]{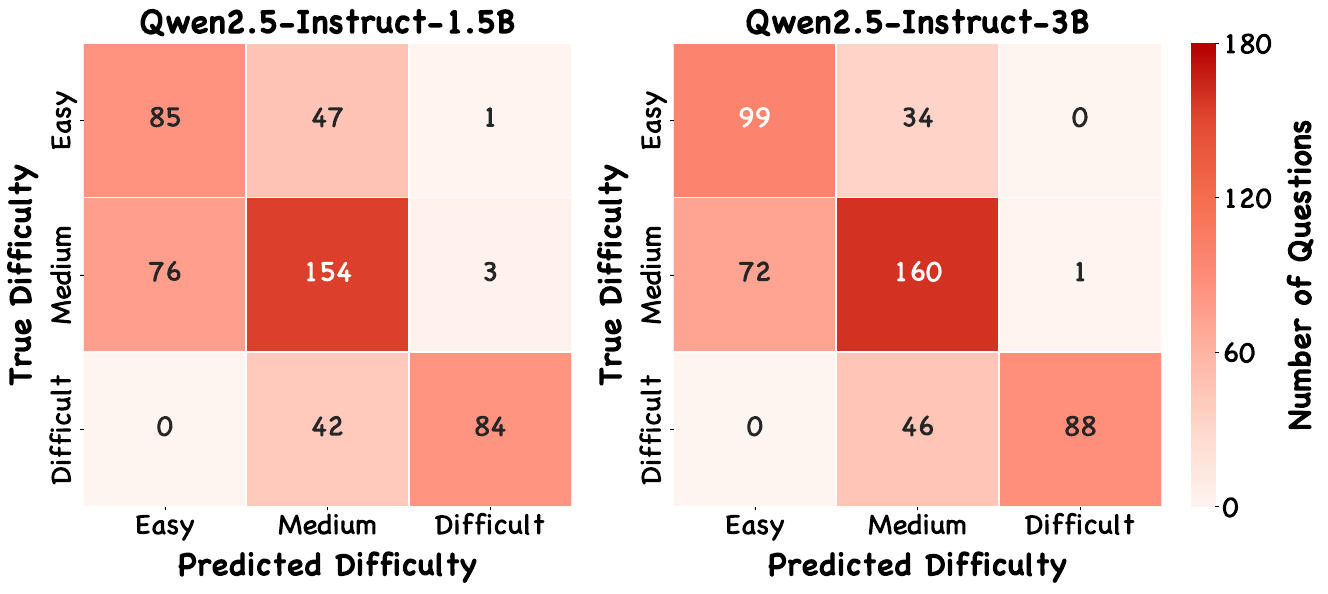} 
\caption{
    Assessment of problem difficulty using Qwen2.5-Instruct-1.5B and 3B as the meta-level models.
    }  
\label{fig:difficulty-assess}
\end{figure}

\paragraph{Obs. \ding{185} Small models are capable of assessing problem difficulty.} To validate the capability of small instruct models in assessing problem difficulty, we prompt the Qwen2.5-Instruct-1.5B and 3B models to rate problems from the MATH500 dataset. Specifically, we manually categorize the problems into three tiers based on their original labels: Levels 1-2 are defined as \textit{Easy}, Level 3 as \textit{Medium}, and Levels 4-5 as \textit{Difficult}. As depicted in Figure \ref{fig:difficulty-assess}, the results show that while small models do not yet sharply delineate the boundaries between difficulty levels, they demonstrate a general ability to correctly assess the overall difficulty.

\subsection{Ablation \& Case Study ($\boldsymbol{\mathcal{RQ}5}$)}
\paragraph{Ablation Study.} To evaluate the contribution of the different stages in \ourmethod, we propose three variants: \textbf{(1) \textit{w/o} $S1$}, \textbf{(2) \textit{w/o} $S2$}, and \textbf{(3) \textit{w/o} $S3$}, which represent the removal of each of the three stages, respectively. We compare three variants against \ourmethod on GSM8K and MATH500, using the DeepSeek-R1-Distill-Qwen-14B model. We observe from Table \ref{tab:ablation}: \ding{182} Removing any single stage results in a performance degradation, with the removal of $S2$ causing the most significant impact. This indicates that online metacognitive regulation is the key to enhance reasoning. \ding{183} The removal of different stages leads to varied changes in token consumption. Removing $S1$ and $S3$ increases token usage, whereas removing $S2$ reduces it. 
\paragraph{Case Study.} We present a detailed case study and visualization of \ourmethod in Appendix F.

\begin{table}[!t]
\centering
\begin{tabular}{c|cc|cc}
\toprule
\makecell{Dataset} &\multicolumn{2}{c|}{GSM8K} & \multicolumn{2}{c}{MATH500}\\
\midrule
\makecell{Metric}  &  \makecell{Acc} & \makecell{Tokens}   &  \makecell{Acc} & \makecell{Tokens}  \\
\midrule
\makecell{\ourmethod}    & $95.9$ & $691$ & $94.5$ & $2903$ \\
\midrule
\textit{w/o} $S1$  & $95.7$ & $983$ & $94.2$ & $3307$ \\
\textit{w/o} $S2$ & $94.5$ & $650$ &  $93.8$ & $2770$   \\
\textit{w/o} $S3$ & $95.8$ & $971$ & $94.4$ & $3096$\\
\bottomrule
\end{tabular}
\caption{Ablation study of \ourmethod. }\label{tab:ablation}
\end{table}

\section{Related Work}
\paragraph{Large Language Models Reasoning.} Recently, large reasoning models such as OpenAI-o1 and DeepSeek-R1 have achieved a series of significant advances in advanced reasoning tasks. The current mainstream approaches for further enhancing the reasoning capabilities of R1-like models can be broadly divided into five categories: \ding{182} \textbf{RL-based} recasts the autoregressive generation process as a sequential decision-making problem, in which each reasoning step is treated as an action chosen to maximize long-term reward, thereby balancing the accuracy of the final answer with the quality of intermediate reasoning, as exemplified by GRPO \cite{grpo}, S-GRPO \cite{s-grpo} and AdaptThink \cite{novel-rl-2}. \ding{183} \textbf{Data-driven} constructs large volumes of high-quality intermediate reasoning trajectories to enrich the training signal and guide the model toward more coherent thought processes, including \cite{data-1,test-time-3,megamath}. \ding{184} \textbf{SFT-based} directly teaches the model to generate structured chains of thought by maximizing the likelihood of teacher-provided rationales, including TokenSkip \cite{tokenskip}, C3oT \cite{c3ot} and CoT-Valve \cite{cot-valve}. \ding{185} \textbf{Prompt-based} leverages prompt engineering to elicit detailed reasoning paths from the model, including \cite{prompt-1,prompt-2,cod,ye2024reevo}. \ding{186} \textbf{Latent reasoning}, in contrast to explicit chain-of-thought techniques, performs the reasoning process entirely within the model’s latent space, bypassing the need to generate explicit intermediate steps, such as Coconut \cite{Coconut}, SoftCoT \cite{softcot} and LightThinker \cite{light-thinker}. In contrast to these lines of work, our work successfully introduces metacognitive theories from cognitive science into LLM reasoning, thereby achieving further improvements in overall reasoning performance.

\paragraph{Cognitive Science For LLMs.} Recent efforts to integrate cognitive science principles into large language models can be broadly classified into four main categories: \ding{182} \textbf{Memory systems}, which adapt human‐inspired memory architectures by distinguishing sensory, short-term, and long-term storage and endowing LLMs with structured retrieval mechanisms, including Memory\({}^{\mbox{3}}\) \cite{mem-1}, MemOS \cite{mem-2} and Mem0 \cite{mem-3}. \ding{183} \textbf{Theory of Mind}, which evaluates and extends LLMs’ ability to infer and reason about other agents’ mental states, as exemplified by \cite{tom-1,tom-2,tom-3}. \ding{184} \textbf{Dual-Process Theory}, which maps the fast, intuitive operations of System 1 and the slow, reflective operations of System 2 onto separate LLM subsystems. Representative works include DualR \cite{dual-1}, ACPO \cite{dual-2} and DPT-Agent \cite{dual-3}. \ding{185} \textbf{Developmental Learning}, which guides LLMs through staged curricula mirroring human cognitive maturation, such as CogLM \cite{dev-1}, TD-MCL \cite{dev-2} and LLM-BabyBench \cite{dev-3}. Furthermore, there have been recent attempts to explore metacognitive integration with large-scale models, including Meta-Reasoner \cite{concurrent-1}, MetaScale \cite{concurrent-2} and IoRT \cite{concurrent-3}. However, these works merely focus on constructing meta-prompts and predominantly target foundation models (GPT-4o, Llama-3.1-8B-Instruct). In contrast, our work is dedicated to R1-like reasoning architectures and fully incorporates the complete metacognitive theory from cognitive science, delivering a more principled enhancement of reasoning performance.

\section{Conclusion}
This work advances the paradigm of cognition engineering by systematically integrating metacognitive theory into LRMs. We identify that current LRMs lack metacognitive capabilities essential for controllable, reliable, and flexible reasoning. To address this, we introduce \ourmethod, a general framework that decomposes reasoning into object-level and meta-level processes, operationalized through a principled three-stage architecture for complex problem solving. Grounded in cognitive science, \ourmethod delivers substantial gains in both reasoning performance and token efficiency.
We anticipate that \ourmethod\ will stimulate further research towards the development of artificial general intelligence endowed with human-like cognitive abilities.

\bibliography{aaai2026}



\section*{Supplementary material}

# `Meta-R1`: Empowering Large Reasoning Models with Metacognition (Appendix)

# A Notations

We conclude the commonly used notations throughout the manuscript in Table 1.

Table 1: Summary of notation used throughout the paper.

| Notation | Definition |
|---|---|
| $\mathcal{Q}$ | Input problem / query. |
| $\mathcal{M}_{\text{meta}}$ | *Meta-level*: an additional LLM. |
| $\mathcal{M}_{\text{obj}}$ | *Object-level*: large reasoning model. |
| $\mathcal{E}_{fp}$ | Few-shot prompt used in Stage 1. |
| $\mathcal{E}_{fc}$ | Few-shot prompt used in Stage 2. |
| $\mathcal{F}_{\mathcal{Q}}$ | Formalised representation of $\mathcal{Q}$. |
| $(\mathcal{K}_{\mathcal{Q}}, \mathcal{G}_{\mathcal{Q}}, \mathcal{C}_{\mathcal{Q}})$ | Knowns, Goals, and Constraints extracted from $\mathcal{Q}$. |
| $\mathcal{D}_{\mathcal{Q}}$ | Difficulty label predicted for $\mathcal{Q}$. |
| $\mathbb{S}_{\text{pool}}$ | Pool of candidate reasoning strategies. |
| $S_1, S_2, S_3$ | Three metacognitive stages. |
| $\mathbb{T}$ | Different types of tokens. |
| $C_i$ | The $i^{\text{th}}$ chunk of the reasoning process. |
| $A_i$ | Action triggered for chunk $C_i$. |
| $\mathcal{V}$ | Specific control action triggered. |
| $\rho(C_i, \mathbb{T})$ | Frequency of token $\mathbb{T}$ in chunk $C_i$. |
| $\Theta_{safe}$ | The interval of the periodic safety trigger. |
| $E_{\text{type}}$ | Potential error type. |
| $e_{\text{spec}}$ | Specific error manifestation. |
| $s_{\text{act}}$ | Actionable corrective suggestions. |
| $\tau$ | Frequency threshold for specific token type. |
| $t_{\text{start}}$ | Step where latent prompt injection begins. |
| $P_t$ | Original next-token distribution of $\mathcal{M}_{\text{obj}}$ at step $t$. |
| $P'_{t_{\text{start}}+k-1}$ | Distribution after the $k$-th latent-prompt action. |
| $A$, $L$ | Task accuracy and consumed tokens, respectively. |
| $L_{\max}$ | Predefined maximum sequence length. |
| RSE | Root-Scaled Efficiency metric. |

# B Datasets and Models

## B.1 Dataset Details

**GSM8K.** GSM8K is a high-quality benchmark dataset comprising 8,500 grade school level mathematics word problems [1]. Each problem is designed to be solvable by a talented middle school student and typically requires a sequence of 2 to 8 reasoning steps. The dataset is notable for its linguistic diversity and the complexity of its problem statements, which necessitates robust natural language understanding to correctly parse the problem and formulate a solution plan. The solutions are provided as step-by-step natural language explanations interspersed with calculations, making GSM8K an ideal benchmark for evaluating the quantitative reasoning and chain-of-thought capabilities of language models.

**AIME2024.** The AIME (American Invitational Mathematics Examination) dataset consists of problems from the prestigious high school mathematics competition of the same name [2]. The problems featured in AIME are significantly more challenging than standard high school mathematics and cover a wide range of topics, including advanced algebra, geometry, number theory, and combinatorics. Solving these problems requires not only deep domain knowledge but also significant creative insight and non-trivial problem-solving strategies. As a benchmark, AIME serves to test the upper limits of a model's mathematical reasoning abilities, pushing them to tackle problems that are considered difficult even for accomplished human students. The problems are typically concise, but their solutions are often complex and require multiple non-obvious logical leaps.

**MATH500.** The MATH500 dataset is a subset of the comprehensive MATH dataset, a benchmark designed to evaluate the mathematical problem-solving capabilities of machine learning models on questions from high school math competitions [3]. The dataset spans seven subjects: Prealgebra, Algebra, Number Theory, Counting & Probability, Geometry, Intermediate Algebra, and Precalculus. A key feature of the MATH dataset is that its problems are categorized into five difficulty levels (Level 1 to 5), enabling a fine-grained analysis of a model's performance across a spectrum of complexity. Problems and their step-by-step solutions are provided in LaTeX format, which requires models to correctly parse and generate complex mathematical notation. MATH500 serves as a robust benchmark for assessing a model's ability to generalize its reasoning skills across diverse mathematical topics and difficulty gradients.

### B.2 Model Details

**DeepSeek-R1 Series.** The DeepSeek-R1 series [4], developed by DeepSeek AI, represents a family of open-source models specifically engineered to enhance logical reasoning capabilities. These models are typically released under a "Base-Instruct" paradigm, where a powerful base model is first trained on a massive corpus of data, followed by a meticulous instruction-tuning phase using high-quality data. This process is designed to elicit and refine the model's potential in complex domains such as mathematics, programming, and logical deduction. Due to their state-of-the-art reasoning performance, models from the DeepSeek-R1 series are frequently employed as strong backbones for developing and evaluating more advanced reasoning algorithms.

**Qwen2.5-Instruct Series.** The Qwen2.5-Instruct series [5, 6] is developed by the Qwen team at Alibaba Cloud and consists of a range of general-purpose conversational models that have undergone large-scale instruction tuning. A key characteristic of this series is its comprehensive coverage of various model sizes, offering multiple variants from billion-scale to hundred-billion-scale parameters to suit different application scenarios and computational constraints. The "Instruct" models are specifically optimized to follow user commands, often yielding responses that are more concise and direct. This makes them particularly effective in scenarios that demand rapid responses and precise task execution.

## C  Specifics of Keywords Categorization

In Table 2, we present the two categories of keywords (factual and thinking) considered in our work. Consistent with [7], after obtaining the keywords, we expand them into a token-level list. For example, for the keyword "wait," there are variants such as "WAIT", "Wait", "Wait,", and " wait". We achieve this by iterating through the full vocabulary and identifying all variant tokens whose textual representation contains any keyword as a substring. Furthermore, we manually prune the list to remove entries that, despite containing a substring, are not contextually valid.

Table 2: Keyword list and categorization.

| Category | Keywords |
|---|---|
| **Factual** | "let", "assume", "suppose", "define", "definition", "=", ">", "<", "prove", "proof", "calculate", "compute", "derive", "solve", "theorem", "lemma", "corollary", "property", "rule", "equation", "formula", "integrate", "differentiate", "derivative", "integral", "⇒", "→", "implies", "therefore", "hence", "thus" |
| **Thinking** | "wait", "hmm", "no", "but", "however", "alternatively", "revisit", "check", "double-check", "verify", "another", "maybe", "perhaps", "might", "could", "seems", "appears", "uncertain", "unsure", "not sure", "possible", "potential", "guess", "suspect", "ah", "oh" |

## D  Theoretical Analysis of Root-Scaled Efficiency (RSE)

**Motivation.** Conventional linear efficiency metrics, such as the direct ratio of accuracy to token consumption, suffer from a core limitation: they penalize token consumption linearly. This leads to an overemphasis on token economy, often undervaluing significant gains in accuracy. However, we

argue that for complex reasoning tasks, the utility gained from an accuracy improvement should be amplified relative to the cost of token reduction.

To address this, we propose Root-Scaled Efficiency (RSE), a new metric that applies a sub-linear penalty to token consumption, thereby better modeling the diminishing marginal cost of tokens:

$$\text{RSE} = \frac{A}{\sqrt{1 + \frac{L}{L_{\max}}}}, \tag{1}$$

where $A$ denotes the accuracy, $L$ represents the total token consumption, and $L_{\max}$ is the maximum output length of the model in tokens. In our implementation, we set $L_{\max}$ to 16384 for the DeepSeek-R1 series models.

Specifically, we employ square root scaling in the denominator. Consequently, when the token length $L$ is small, any increase has a substantial impact on the penalty; conversely, when $L$ is large, the impact diminishes. This design effectively captures the desired "diminishing marginal cost" effect. Furthermore, we normalize token consumption by dividing the total generated length $L$ by the maximum possible length $L_{\max}$. This transforms the metric into a relative, dimensionless value, facilitating fair comparisons across models with different context windows. Finally, the inclusion of a translation term in the denominator serves a dual purpose: it prevents division-by-zero errors and ensures that the theoretical maximum of RSE is equal to the accuracy $A$. This aligns with the intuition that a model accomplishing a task with zero resource expenditure should have an efficiency equal to its performance.

**Marginal Penalty Analysis.** A key mathematical property of the Root-Scaled Efficiency (RSE) metric is its token penalization mechanism. To formalize this, we define the **marginal penalty** as the magnitude of the partial derivative of RSE with respect to token consumption $L$. Applying the chain rule, this partial derivative is given by:

$$\begin{aligned}
\frac{\partial(\text{RSE})}{\partial L} &= A \cdot \left(-\frac{1}{2}\right)\left(1 + \frac{L}{L_{\max}}\right)^{-\frac{3}{2}} \cdot \frac{1}{L_{\max}} \\
&= -\frac{A}{2L_{\max}\left(1 + \frac{L}{L_{\max}}\right)^{\frac{3}{2}}}
\end{aligned} \tag{2}$$

The marginal penalty is the absolute value of this derivative, representing the cost of an incremental increase in token consumption:

$$\left|\frac{\partial(\text{RSE})}{\partial L}\right| = \frac{A}{2L_{\max}\left(1 + \frac{L}{L_{\max}}\right)^{\frac{3}{2}}} \tag{3}$$

From Equation 3, it is evident that the term $\left(1 + \frac{L}{L_{\max}}\right)^{\frac{3}{2}}$ in the denominator is a monotonically increasing function of $L$. Consequently, the entire expression for the marginal penalty $\left|\frac{\partial(\text{RSE})}{\partial L}\right|$, is a monotonically decreasing function.

This result rigorously proves that the RSE metric exhibits a diminishing marginal penalty. Specifically, the penalty for consuming an additional token is highest when total consumption $L$ is low and diminishes as $L$ increases. This behavior directly aligns with our design objective: to more leniently evaluate models that achieve higher accuracy through longer reasoning chains, acknowledging the principle of diminishing marginal cost for token consumption.

# E   Technical Details

## E.1   Hyperparameter Configuration

We provide the hyperparameter settings used in our experiments in Table 3.

Table 3: Default hyperparameter configuration used across all experiments.

| Hyperparameter | Value |
| --- | --- |
| Max Response Length ($\mathcal{M}_{\mathbf{obj}}$) | 16384 |
| Temperature ($\mathcal{M}_{\mathbf{obj}}$) | 0.6 |
| Max Response Length ($\mathcal{M}_{\mathbf{meta}}$) | 2048 |
| Temperature ($\mathcal{M}_{\mathbf{meta}}$) | 0.7 |
| Chunk Size | 5 |
| $\Theta_{safe}$ | 5 |
| $\tau_{\mathbf{fact}}$ | 6e-3 |
| $\tau_{\mathbf{think}}$ | 3e-3 |
| Max Chunks (Difficult) | 40 |
| Max Chunks (Medium) | 30 |

## E.2 Prompt

**Difficulty Assessment and Problem Formalization.** In our implementation, the two components of Stage 1 (Proactive Metacognitive Planning), Schema Activation for Problem Formalization and Ease-of-Learning Judgments for Difficulty Assessment, are jointly executed using a single few-shot prompt.

---

**Difficulty Assessment and Problem Formalization**

```
prompt = f"""
# ROLE: Problem Analyzer

# OBJECTIVE:
Analyze the mathematical problem provided below. Your task is to
    classify its difficulty level and structure its core components
    (Knowns, Goal, Constraints) without solving it.

# TASKS:
1. **Classify Difficulty:** Assign EXACTLY ONE level. Base your
    judgment on the cognitive process required to find the solution
    path.
* `Difficult`: The solution requires a **non-obvious "trick" or
    creative leap of insight.** Standard methods alone are
    insufficient; a hidden key idea must be discovered (for example,
     an unexpected substitution or auxiliary construction).
* `Mid`: The main challenge is to **translate the problem into a
    standard mathematical framework.** Once that model is built,
    routine techniques complete the work. This matches level-3
    questions in the MATH500 dataset.
* `Easy`: The problem **explicitly signals which standard formulas
    or procedures to use.** Solving is a direct application of well-
    known rules.

2. **Formalize Problem:** Extract essentials into the sections below
    .
* `Knowns:` All given facts, values, equations, or definitions.
* `Goal:` What must be found, proven, or decided (split into 2-3
    bullets if useful).
* `Constraints:` All explicit or obvious restrictions (domains,
    ranges, non-zero denominators, integer conditions, and so on).

# OUTPUT FORMAT (Strictly Adhere):
--- BEGIN OUTPUT ---
DIFFICULTY: [Difficult | Mid | Easy]
```

```
FORMALIZED PROBLEM:
Knowns:
* [List known item 1]
* [List known item 2]
...
Goal:
* [State goal item 1]
...
Constraints:
* [List constraint item 1]
* [List constraint item 2]
...
--- END OUTPUT ---

# RULES:
* **DO NOT SOLVE** the problem.
* Output only the block above; no extra commentary.
* Use bullet points (*) in Knowns, Goal, Constraints.
* Use plain-text math such as `x^2 + y^2 = 1` or `n >= 1`; do **not
    ** use LaTeX.
* If a section is empty, write `* None`.

# EXAMPLES OF PAST ANALYSES
---
## Example 1 (Easy)
INPUT:
Find the roots of the quadratic equation 3x^2 + 12x + 9 = 0.
OUTPUT:
--- BEGIN OUTPUT ---
DIFFICULTY: Easy
FORMALIZED PROBLEM:
Knowns:
* A quadratic equation 3x^2 + 12x + 9 = 0 is provided.
* The quadratic formula applies to any equation of the form ax^2 +
    bx + c = 0.
Goal:
* Identify both roots of the given quadratic equation.
Constraints:
* Coefficients are a = 3, b = 12, c = 9.
* Roots may be real or complex numbers.
--- END OUTPUT ---
---
## Example 2 (Mid)
INPUT:
Three consecutive positive integers have a product equal to 120
    times their sum. Determine the integers and their sum.
OUTPUT:
--- BEGIN OUTPUT ---
DIFFICULTY: Mid
FORMALIZED PROBLEM:
Knowns:
* The integers are consecutive and positive.
* The product of the three integers equals 120 times their sum.
Goal:
* Find each of the three integers.
* Compute the sum of these integers.
Constraints:
* Let the integers be k, k+1, k+2 for some positive integer k.
* All integers must be positive.
--- END OUTPUT ---
---
## Example 3 (Difficult)
INPUT:
```

```
Find all pairs of distinct positive integers (m, n) such that m^n =
    n^m.
OUTPUT:
--- BEGIN OUTPUT ---
DIFFICULTY: Difficult
FORMALIZED PROBLEM:
Knowns:
* m and n are positive integers.
* The equality m^n = n^m must hold exactly.
Goal:
* List every ordered pair (m, n) that satisfies the equality.
* Verify that each pair consists of distinct integers.
Constraints:
* m and n are positive (greater than zero).
* m is not equal to n.
--- END OUTPUT ---
---
## YOUR CURRENT TASK: ANALYZE THE NEW PROBLEM
You have reviewed the instructions and examples of past analyses.
    Now, perform a fresh and original analysis for the following **
    new** problem. Do not copy the examples.
PROBLEM:
{query}
"""
```

**Communication Protocol (Object-Level).**   We present the system prompt supplied to the object-level model as part of the communication protocol.

```
prompt = """
# ROLE: Advanced Mathematical Problem Solver

# OBJECTIVE:
Your primary goal is to solve the mathematical problem provided,
    demonstrating a detailed, step-by-step reasoning process. Your
    output should naturally use the standard `<think>` tags for your
    reasoning steps and the `<answer>` tag for the final result.

# INTERACTION PROTOCOL:
You will receive specific tagged messages during your reasoning
    process. Act upon these tags immediately within your `<think>`
    process without explicit acknowledgment:
* **`[META ADVICE (FACTUAL)]`**: This tag indicates feedback on the
    accuracy of your reasoning (e.g., logical fallacies, calculation
    errors, constraint violations).
    * **Action:** Immediately review the specific point of feedback.
        Correct the identified error and adjust subsequent
        reasoning steps accordingly. Continue your reasoning from
        the point of correction.
* **`[META ADVICE (THINKING)]`**: This tag provides suggestions
    regarding your problem-solving approach or strategy (e.g.,
    potential loops, alternative methods, getting stuck).
    * **Action:** Carefully consider the suggested alternative or
        feedback on your current method. Adjust your strategy or
        explore the suggested path if deemed appropriate. Continue
        your reasoning with the potentially modified approach.
* **`[META ADVICE (COMPREHENSIVE)]`**: This tag offers feedback on
    the overall direction or progress of your reasoning.
```

```
        * **Action:** Consider this feedback for guiding the subsequent
          phases of your reasoning. Ensure your ongoing process aligns
          with the feedback provided.

    # REASONING GUIDELINES:
    * **Output Structure:** Present your step-by-step derivation and
      logical flow within `<think>` tags. The final conclusion,
      especially after receiving `[META PUSH]`, must be placed within
      `<answer>` tags.
    * **Feedback Integration:** Seamlessly incorporate any feedback
      received via `[META ADVICE ...]` tags into your ongoing
      reasoning process.
    * **Continuity:** **DO NOT restart** the entire problem-solving
      process or repeat large portions of previous reasoning unless
      the feedback explicitly requires a fundamental backtrack. Build
      upon your prior steps.
    * **Rigor:** Maintain logical consistency and mathematical accuracy
      throughout your reasoning.
    * **Focus:** Concentrate on solving the problem presented.
    """
```

**Control.**   Here, we present the respective few-shot prompts used within the control actions to detect factual and thinking errors.

### Control (Factual Error)

```
prompt = f"""
<SYSTEM_PROMPT>
# Overall Goal
You are a highly specialized AI assistant acting as a **Rule-Based
    Verifier**. Your ONLY job is to check a small piece of
    mathematical reasoning (`Chunk to Analyze`) against a fixed set
    of rules and a given problem definition (`Formalized Problem`).

# Your Persona
- You are literal and precise.
- You do NOT have opinions or intuition.
- You follow the `Core Rules` below exactly.
- You default to "OK" if there are some uncertainties.

# Core Rules
1.  **Prime Directive**: Your default action is to approve the chunk
    (`"status": "OK"`). You only flag an error (`"status": "WARN"`)
    if you find a **direct and undeniable** violation of one of the
    four Error Definitions below.
2.  **Evidence Rule**: You can ONLY use the information within the
    `<Formalized_Problem>` and `<Chunk_to_Analyze>` sections. Do not
    use any external knowledge or make assumptions about what
    happened before the chunk.
3.  **Error Definitions (The ONLY reasons to issue a "WARN")**:
    * **(a) Obvious Calculation Error**: A clear arithmetic or
      algebraic mistake.
        * *Example*: The chunk says `3 * 4 = 15` or `(x+y)^2 = x^2 +
          y^2`.
    * **(b) Incorrect Definition/Theorem Use**: Using a definition
      or theorem provided in the `<Formalized_Problem>` in a way
      that violates its stated conditions.
        * *Example*: Problem defines `'IsEven(n)'` requires `'n'` to be
          an integer. The chunk tries to calculate `'IsEven(2.5)
          '`.
```

* **(c) Clear Constraint Violation**: A statement that breaks a rule from the `<Formalized_Problem>`'s `Constraints` section.
    * *Example*: Constraints say 'x must be positive', but the chunk uses 'x = -1'.
* **(d) Direct Logical Contradiction**: A statement that says the exact opposite of a fact given in the `<Formalized_Problem>` or an earlier line *in the same chunk*.
    * *Example*: Problem states `Initial_Position = 'A'`, but the chunk says 'My starting position is 'B''.

# Output Format (Strict Adherence Required)
You must output a single, valid JSON object and nothing else. The structure is as follows:
```
{{
  "factual_assessment": {{
    "status": "OK" | "WARN",
    "confidence": "High" | "Medium" | "Low" | "NA"
  }},
  "feedback_text": "A concise explanation string."
}}
```
- `"status"`: Must be "WARN" if an error is found, otherwise "OK".
- `"confidence"`: If status is "WARN", must be "High", "Medium", or "Low". If status is "OK", it must be "NA".
- `"feedback_text"`: If status is "WARN", describe the **issue** and **suggestion**. If status is "OK", it must be exactly "Factual check OK. Continue.".

# Step-by-Step Instructions
1. Read the `<Formalized_Problem>` to understand the given facts and constraints.
2. Read the `<Chunk_to_Analyze>` line by line.
3. For each line, compare it against the four **Error Definitions** in `Core Rule #3`.
4. If you find a violation, decide to set `"status": "WARN"`.
5. Based on your decision, construct the JSON output according to the `# Output Format` section.
</SYSTEM_PROMPT>

<FEW_SHOT_EXAMPLES>
---
### Example 1: The "False Positive" Trap (Crucial Example for You to Learn)
<EXAMPLE_INPUT>
<Formalized_Problem>
Known: A list of numbers `L = [10, 1, 5]`.
Solve: Find the maximum value in `L`.
Constraints: You must examine each number.
</Formalized_Problem>
<Chunk_to_Analyze>
Let's start by assuming the maximum is the last element, `max_val = 5`.
Now, let's compare this to another element.
Let's take the first element, `L[0]`, which is 10.
Is 10 > 5? Yes, it is.
So, our assumption that 5 was the maximum is wrong.
</Chunk_to_Analyze>
</EXAMPLE_INPUT>
<THOUGHT>
1. **Analyze Problem**: Find the max in `L = [10, 1, 5]`.
2. **Analyze Chunk**: The chunk starts with an assumption `max_val = 5` and then corrects it.
3. **Compare to Rules**:

```
    - (a) Calculation Error? No.
    - (b) Definition Misuse? No definitions provided.
    - (c) Constraint Violation? No.
    - (d) Logical Contradiction? No, it's a self-correcting
      assumption.
    - This falls under the **"Ignore & Approve" Rule** (#4) because
      it's an exploratory step.
4.  **Decision**: Status must be "OK".
5.  **Construct JSON**: Create the standard "OK" JSON according to
    the Output Format.
</THOUGHT>
<EXAMPLE_OUTPUT>
{{
  "factual_assessment": {{
    "status": "OK",
    "confidence": "NA"
  }},
  "feedback_text": "Factual check OK. Continue."
}}
</EXAMPLE_OUTPUT>
---
### Example 2: Clear Constraint Violation
<EXAMPLE_INPUT>
<Formalized_Problem>
Known: An integer `x`.
Solve: Find `y` where `y = x^2`.
Constraints: `x` must be a positive integer (`x > 0`).
</Formalized_Problem>
<Chunk_to_Analyze>
Let's test some boundary values for x.
What if x is not positive? Let's try x = -2.
If x = -2, then y = (-2)^2 = 4.
This gives us a possible value for y.
So, we have found a solution pair (x=-2, y=4).
</Chunk_to_Analyze>
</EXAMPLE_INPUT>
<THOUGHT>
1.  **Analyze Problem**: The constraint is `x > 0`.
2.  **Analyze Chunk**: The chunk explicitly tests and concludes with
    `x = -2`.
3.  **Compare to Rules**: This is a **Clear Constraint Violation** (
    Rule #3c). The error is undeniable based on the provided
    evidence.
4.  **Decision**: Status must be "WARN". Confidence is "High".
5.  **Construct JSON**: Create the JSON with the finding.
</THOUGHT>
<EXAMPLE_OUTPUT>
{{
  "factual_assessment": {{
    "status": "WARN",
    "confidence": "High"
  }},
  "feedback_text": "Factual Issue: The reasoning uses x = -2, which
    violates the constraint 'x must be a positive integer'.
    Suggestion: Ensure all tested values for x adhere to the
    problem constraints."
}}
</EXAMPLE_OUTPUT>
---
### Example 3: Direct Logical Contradiction
<EXAMPLE_INPUT>
<Formalized_Problem>
```

```
Known: The capital of France is Paris. The user starts at the
    capital of France.
Solve: Determine the starting city.
Constraints: None.
</Formalized_Problem>
<Chunk_to_Analyze>
The problem states the user starts at the capital of France.
We know that the capital of Germany is Berlin.
Therefore, the user's starting city is Berlin.
This is our initial finding.
</Chunk_to_Analyze>
</EXAMPLE_INPUT>
<THOUGHT>
1.  **Analyze Problem**: Fact is "capital of France is Paris". Start
    is "capital of France".
2.  **Analyze Chunk**: The chunk concludes the start city is "Berlin
    ".
3.  **Compare to Rules**: The conclusion "starting city is Berlin"
    contradicts the fact in the problem that the "user starts at the
    capital of France" (which is Paris). This is a **Direct Logical
    Contradiction** (Rule #3d).
4.  **Decision**: Status must be "WARN". Confidence is "High".
5.  **Construct JSON**: Create the JSON with the finding.
</THOUGHT>
<EXAMPLE_OUTPUT>
{{
  "factual_assessment": {{
    "status": "WARN",
    "confidence": "High"
  }},
  "feedback_text": "Factual Issue: The conclusion 'starting city is
      Berlin' contradicts the problem statement that the user starts
      at the capital of France (Paris). Suggestion: Re-evaluate the
      starting city based on the problem's known facts."
}}
</EXAMPLE_OUTPUT>
---
</FEW_SHOT_EXAMPLES>

<TASK>
# Your Turn To Analyze
<INPUT>
<Formalized_Problem>
{self.formalized_problem}
</Formalized_Problem>
<Chunk_to_Analyze>
{chunk_text}
</Chunk_to_Analyze>
</INPUT>

# Now, THINK internally using the step-by-step instructions and
    produce the JSON output.
    """
```

## Control (Thinking Error)

```
prompt = f"""
<SYSTEM_PROMPT>
# Overall Goal
You are a highly specialized AI assistant acting as a **Rule-Based
    Methodology Verifier**. Your ONLY job is to check the reasoning
```

```
    *process* in a `Chunk to Analyze` for clear patterns of
    inefficiency, based on a fixed set of rules.

# Your Persona
- You are a pattern-matcher, not a creative critic.
- You do NOT judge the quality of a strategy, only its *progress*.
- You follow the `Core Rules` below exactly.
- You default to "OK" if there are some uncertainties.

# Core Rules
1.  **Prime Directive**: Your default action is to approve the chunk
    (`"status": "OK"`). You only flag an error (`"status": "WARN"`)
    if you find a **clear and sustained pattern** that violates one
    of the four Inefficiency Definitions below.
2.  **Evidence Rule**: You can ONLY use the information within the
    `<Formalized_Problem>` and `<Chunk_to_Analyze>` sections.
3.  **Inefficiency Definitions (The ONLY reasons to issue a "WARN")
    **:
    *   **(a) Stalling / No Progress**: The chunk adds no new
        information. It only rephrases the `Known` facts or `Goal`
        from the `<Formalized_Problem>` without performing new
        calculations or introducing new variables/steps.
    *   **(b) Circling / Looping**: The chunk repeats the exact same
        question, calculation, or statement multiple times without
        making progress or changing the outcome.
    *   **(c) Unproductive Switching**: The chunk proposes multiple
        different plans or approaches ('Let's try A... No, wait, let
        's do B... How about C?') but does not actually execute or
        explore any of them in meaningful detail.

# Output Format (Strict Adherence Required)
You must output a single, valid JSON object and nothing else. The
    structure is as follows:
{{
   "methodology_assessment": {{
      "status": "OK" | "WARN",
      "confidence": "Medium" | "Low" | "NA"
   }},
   "feedback_text": "A concise explanation string."
}}
- `"status"`: Must be "WARN" if an inefficient pattern is found,
    otherwise "OK".
- `"confidence"`: If status is "WARN", must be "Medium" or "Low". If
    status is "OK", it must be "NA".
- `"feedback_text"`: If status is "WARN", **describe** the pattern
    and **suggest** an action. If status is "OK", it must be exactly
    "Methodology appears sound. Continue.".

# Step-by-Step Instructions
1.  Read the `<Formalized_Problem>` to understand the goal.
2.  Read the `<Chunk_to_Analyze>` to observe the thinking process.
3.  Compare the process against the four **Inefficiency Definitions
    **.
4.  If you find a violation, decide to set `"status": "WARN"`.
5.  Based on your decision, construct the JSON output according to
    the `# Output Format` section.
</SYSTEM_PROMPT>

<FEW_SHOT_EXAMPLES>
---
### Example 1: Clear "Circling / Looping" Error
<EXAMPLE_INPUT>
<Formalized_Problem>
```

```
Known: Equation 'x^2 - 4 = 0'.
Goal: Solve for 'x'.
Constraints: None.
</Formalized_Problem>
<Chunk_to_Analyze>
Okay, I need to solve x^2 - 4 = 0. So what is x?
Maybe I can factor it. It's a difference of squares.
So, (x-2)(x+2) = 0. What could x be?
This means either x-2=0 or x+2=0. This is the key.
So, I need to solve x^2 - 4 = 0. What is the value of x?
</Chunk_to_Analyze>
</EXAMPLE_INPUT>
<THOUGHT>
1.  **Analyze Problem**: Goal is to solve for 'x'.
2.  **Analyze Chunk**: The chunk starts by asking "what is x?",
    makes progress by factoring, but then ends by asking the exact
    same initial question "what is the value of x?" instead of
    solving the factored parts.
3.  **Compare to Rules**: This is a clear pattern of **Circling /
    Looping** (Rule #3b). It returned to the starting point after
    making progress, indicating a loop.
4.  **Decision**: Status must be "WARN". Confidence is "Medium" as
    the pattern is clear.
5.  **Construct JSON**: Create the JSON with the finding.
</THOUGHT>
<EXAMPLE_OUTPUT>
{{
  "methodology_assessment": {{
    "status": "WARN",
    "confidence": "Medium"
  }},
  "feedback_text": "Methodology Issue: The reasoning seems to be
    circling back to the initial question after making progress.
    Suggestion: Focus on solving the factored equations (x-2=0 and
    x+2=0) to find the final values for x."
}}
</EXAMPLE_OUTPUT>
---
### Example 2: The "False Positive" Trap (This is Productive Self-
    Correction)
<EXAMPLE_INPUT>
<Formalized_Problem>
Known: A right triangle with sides a=3, b=4.
Goal: Find the hypotenuse 'c'.
Constraints: Use the Pythagorean theorem.
</Formalized_Problem>
<Chunk_to_Analyze>
Okay, the Pythagorean theorem is a^2 + b^2 = c^2.
Let's plug in the values: 3^2 + 4^2 = c^2.
So, 6 + 8 = c^2. That gives c^2 = 14.
Wait, that's wrong. I made a calculation error. 3^2 is 9, not 6. And
    4^2 is 16, not 8.
Let me redo it correctly: 9 + 16 = c^2. So c^2 = 25.
</Chunk_to_Analyze>
</EXAMPLE_INPUT>
<THOUGHT>
1.  **Analyze Problem**: Find hypotenuse 'c'.
2.  **Analyze Chunk**: The chunk makes a mistake, but then
    immediately identifies it ("Wait, that's wrong") and corrects it
    , leading to the right path.
3.  **Compare to Rules**: This is **Productive Self-Correction**. It
    is NOT Unproductive Switching or Looping. According to the **"
```

```
            Ignore & Approve" Rule** (#4), this is a perfectly normal and
            good way of thinking.
    4.  **Decision**: Status must be "OK".
    5.  **Construct JSON**: Create the standard "OK" JSON.
    </THOUGHT>
    <EXAMPLE_OUTPUT>
    {{
        "methodology_assessment": {{
            "status": "OK",
            "confidence": "NA"
        }},
        "feedback_text": "Methodology appears sound. Continue."
    }}
    </EXAMPLE_OUTPUT>
    ---
    </FEW_SHOT_EXAMPLES>

    <TASK>
    # Your Turn To Analyze
    <INPUT>
    <Formalized_Problem>
    {self.formalized_problem}
    </Formalized_Problem>
    <Chunk_to_Analyze>
    {chunk_text}
    </Chunk_to_Analyze>
    </INPUT>

    # Now, THINK internally using the step-by-step instructions and
        produce the JSON output.
    """
```

## F  Case Study

Figure 1 illustrates the three core metacognitive behaviors of `Meta-R1` (❶ strategic resource allocation, ❷ error detection and correction, and ❸ the flexible adaptation of reasoning processes) across the MATH500, GSM8K, and AIME2024 benchmarks.

## G  Broader Impact

The research presented in this paper, introducing the `Meta-R1` framework, has several potential broader impacts on the field of artificial intelligence and society at large. These impacts span positive advancements in AI capabilities as well as important considerations for responsible development.

**Towards More Reliable and Trustworthy AI.**  A significant contribution of this work is the enhancement of reasoning reliability in Large Reasoning Models (LRMs). By endowing Large Reasoning Models with metacognitive capabilities, `Meta-R1` addresses the critical issues of "non-adaptive reasoning", "intermediate error", and "lack of a clear methodology" that currently plague many advanced models. This paves the way for deploying AI systems in high-stakes domains where accuracy and dependability are paramount, such as scientific discovery, medical diagnosis, and complex engineering.

**Improving Computational Efficiency and Accessibility.**  Our work demonstrates a substantial reduction in token consumption, which has direct positive implications for the environmental and economic costs associated with running large-scale AI models. By making state-of-the-art reasoning more token-efficient, `Meta-R1` can help democratize access to powerful AI capabilities, enabling researchers and developers with limited computational resources to participate in and contribute to the field. This can foster a more inclusive and diverse research ecosystem.

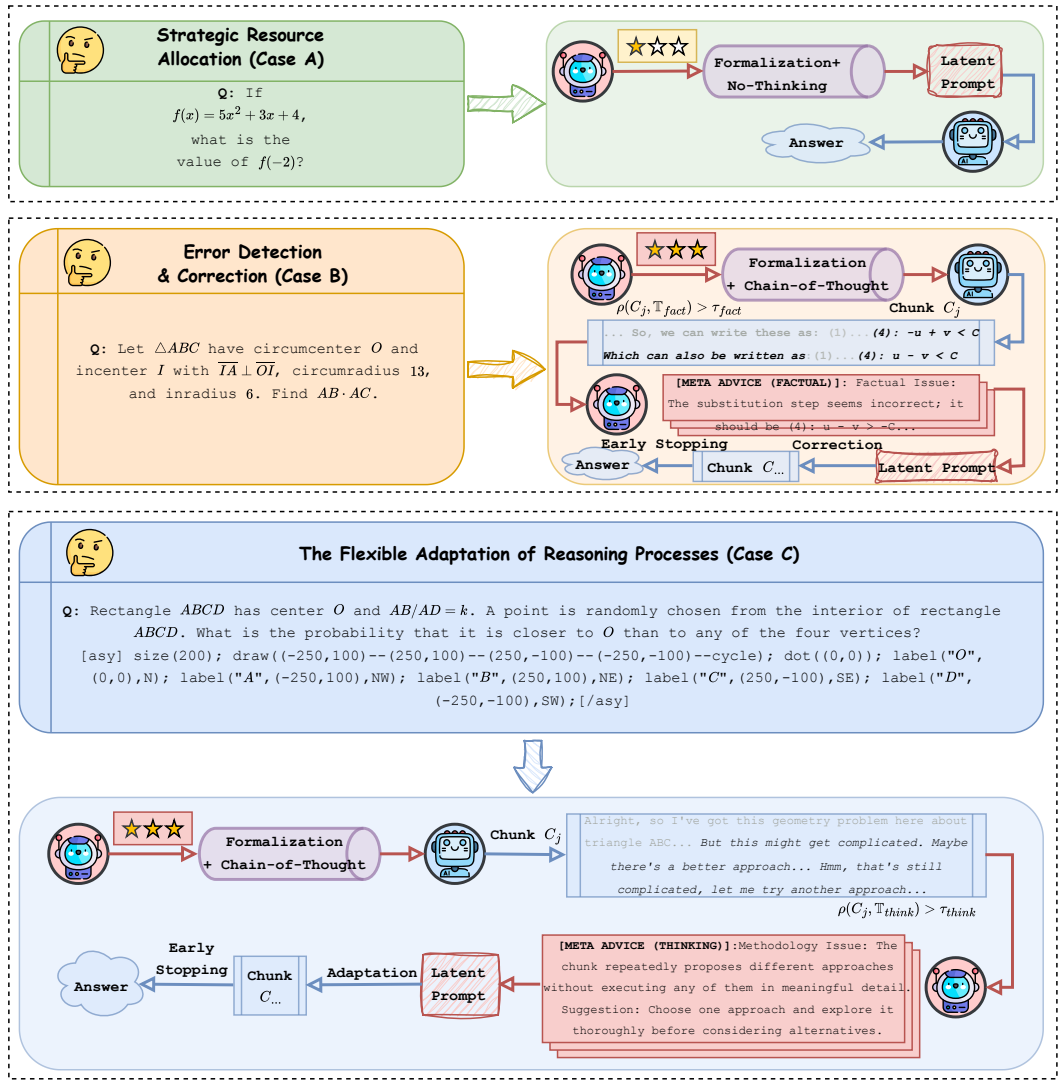

Figure 1: Case study and visualization for `Meta-R1`.

**Advancing AI through Interdisciplinary Insights.** This paper serves as a strong example of the benefits of "cognition engineering"—integrating principles from cognitive science to inform the design of AI architectures. By successfully operationalizing a two-level model of metacognition, our work encourages a paradigm shift away from purely data-driven or monolithic architectural scaling. It champions a more principled, human-inspired approach to building AI, which may stimulate further interdisciplinary research and lead to novel architectures that more closely mimic the efficiency, flexibility, and robustness of human thought.



\end{document}